\documentclass{article}
\usepackage{fullpage}
\usepackage[utf8]{inputenc}
\usepackage[T1]{fontenc}
\usepackage{amssymb}
\usepackage{booktabs}
\usepackage{longtable}
\usepackage{multirow}
\usepackage{amsmath}
\usepackage[hyphens]{url}
\usepackage{color}
\usepackage{graphicx}
\usepackage{soul}
\usepackage{arydshln}
\usepackage[textwidth=3cm,textsize=tiny]{todonotes}
\usepackage{natbib}

\title{An Analysis of the Semantic Annotation Task on the \\ Linked Data Cloud}

\date{}

\author{{\bf Michel Gagnon}\\
Department of computer engineering and software engineering, \\ 
Polytechnique Montréal \\
E-mail: michel.gagnon@polymtl.ca\\ \\
{\bf Amal Zouaq} \\
Department of computer engineering and software engineering, \\ 
Polytechnique Montréal \\
School of Electrical Engineering and Computer Science, University of Ottawa\\
E-mail: amal.zouaq@polymtl.ca\\ \\
{\bf Francisco Aranha} \\
Fundação Getulio Vargas,\\ Escola de Administração de Empresas de São Paulo \\  \\
{\bf Faezeh Ensan} \\
Ferdowsi University of Mashhad, Mashhad Iran, \\ 
 E-mail: ensan@um.ac.ir\\ \\
 {\bf Ludovic Jean-Louis}\\
Netmail \\ E-mail: ludovic.jean-louis@netmail.com}

\begin{document}

\maketitle

\begin{abstract}
Semantic annotation, the process of identifying key-phrases in texts and linking them to concepts in a knowledge base, is an important basis for semantic information retrieval and the Semantic Web uptake. Despite the  emergence of semantic annotation systems, very few comparative studies have been published on their performance. In this paper, we provide an evaluation of the performance of existing systems over three tasks: full semantic annotation, named entity recognition, and keyword detection. More specifically, the spotting capability (recognition of relevant surface forms in text) is evaluated for all three tasks, whereas the disambiguation (correctly associating an entity from Wikipedia or DBpedia to the spotted surface forms) is evaluated only for the first two tasks. Our evaluation is twofold:  First, we compute standard precision and recall on the output of semantic annotators on diverse datasets, each best suited for one of the identified tasks. Second, we build a statistical model using logistic regression to identify significant performance differences. Our results show that systems that provide full annotation perform better than named entities annotators and keyword extractors, for all three tasks. However, there is still much room for improvement for the identification of the most relevant entities described in a text.
\end{abstract}


\section{Introduction}
\label{section:intro}

Semantic annotation is an important basis for realizing the Semantic Web vision \citep{dill2003semtag,Shen2015}, a vision of a Web of machine-understandable data, and an important foundation for retrieving semantic information. Semantic annotation involves the recognition of short text fragments called \textit{mentions} in documents (aka \textit{spotting}) and links them to URIs defined in a knowledge base, aka \textit{disambiguation}. Originally, automatic semantic annotation has been implemented using well-defined and restricted ontologies and knowledge bases \citep{kiryakov2011}. This led to several platform such as KIM \citep{kiryakov2011} or Apache Stanbol \citep{sinaci2012semantic}. The emergence of the linked data cloud has encouraged the development of several annotation services \citep{Milne2013222,Ferragina:2010,Mendes2011}  such as DBpedia Spotlight and Yahoo which exploit LOD datasets and especially DBpedia/Wikipedia \citep{bizer2009dbpedia} as their background knowledge bases.  These knowledge bases, with their wide coverage, their structured description of content and their dynamic nature, are well-suited for enriching almost all types of unstructured text. However, they also raise new challenges due to their size and their cross-domain nature. Thus, it is not surprising that, among the various services that appeared in the last few years, we see a great variation in terms of performance \citep{cornolti2013framework,PRICAI2014,Chen2013,Ruiz2015,derczynski2015,gangemi2013}. 

Additionally, mentions in text might represent entities, concepts, keywords, multi-word expressions, events, etc. and depending on the task at hand, some types of mentions might be more appropriate. While the majority of linked data annotators are described as "Semantic Annotators" without any specific type of mention in mind, in practice, many are more geared towards named entities (e.g. organizations, people) than topics or keywords (e.g. Artificial intelligence) for instance. It is thus often difficult to distinguish the most adequate service among the plethora of available Web APIs. In this paper, our aim is to facilitate such a choice, formalize semantic annotation tasks as well as assess some of the available linked data semantic annotators’ strengths and weaknesses for these tasks. Note that this paper does not aim at providing an exhaustive survey of existing annotation APIs but rather focuses on some prominent APIs and describes a methodology for the evaluation of Semantic annotators.

Based on our analysis of the state of the art, we identified that semantic annotators can be applied to three main tasks: 
\begin{description}
\item[Traditional semantic annotation (SA):] Given a particular knowledge base, SA consists in the identification of all the possible KB entities in a document. Here mentions can represent keywords, classes, individuals and might be of any type. The early semantic annotation platform KIM \citep{kiryakov2011} is a good example of such an approach, which is often based on the assumption of a closed knowledge base. In linked data-based semantic annotators, mostly those based on DBpedia/Wikipedia, all Wikipedia content (aka resources) can be identified in documents.  
\item[Named entity annotation (NE):] The second task focuses on the annotation of named entities, which refer to individuals of certain types. Named entity annotation is an extension of the simpler task of named entity recognition (NER), an important topic in natural language processing that has been vastly studied and investigated in the literature \citep{nadeau2007survey}. The main difference is that traditional NER has very limited types such as PERSON and ORGANIZATION which are generally not defined in an ontology.  On top of these traditional named entities, current linked data-based annotators define an extended range of named entities and rely on a finer classification of each named entity (e.g. politicians, poets and non-governmental organizations).

\item[Keyword extraction (KW):] The third task can be described as the identification of a limited number of \textit{prominent} domain-related key-phrases and concepts. An example would be the extraction of key-phrases related to a specific research topic in academic publications \citep{qureshi2012short} or the identification of biologically significant phrases related to protein functions \citep{andrade1998automatic}. This task requires filtering and ranking capabilities that identify the most important mentions. Compared to traditional keyword extractors, linked data semantic annotators can also (not always) link the extracted keywords to their corresponding concepts in a knowledge base.

\end{description}
 
 These three tasks are used as a basis for evaluating and predicting the performance of some of the most prominent semantic annotators on similar datasets. While there are frameworks such as GERBIL \citep{gerbil2015} which handle the evaluation of semantic annotators, none has made the distinction based on the three tasks as described above. Additionally, our results show that it does not suffice to compare metrics' results to evaluate the interest of semantic annotators. As we will see in this paper, a finer statistical analysis indicates that some semantic annotators' results are indistinguishable. 

The remainder of this paper is organized as follows: in Section \ref{section:state-of-art}, we briefly present the limited state of the art on semantic annotators' evaluation. Section \ref{section:methodology} describes our research methodology, including our research question, datasets and  evaluation metrics. We also provide a description of the evaluated annotators. The following three sections describe our experimental results, first by taking macro-averages over documents, then by considering annotations independently from documents (micro-average) and finally by considering only the three categories of systems instead of their individual performances. Section \ref{section:discussion} discusses our findings and the limitations of this study, and concludes with further discussion on the evaluation strategies and results.

\section{State of the Art}
\label{section:state-of-art}
Due to the recent development of semantic annotation systems, very few comparative studies have been published on their performance \citep{www2013,joksimovic2013empirical}, especially in the three tasks mentioned above. Existing evaluation results are mostly related to specific semantic annotation services (e.g. \cite{Mendes2011} and \cite{ferragina2010tagme}), and hence are based on diverse metrics and gold standards, different data gathering methodologies and a limited set of evaluation datasets. In general, these works do not include numerous annotation systems for their evaluation and comparison. Two significant exceptions are the works reported by \cite{www2013} and \cite{gerbil2015}. In these works, the authors provide a framework for benchmarking semantic annotation systems and comparing their performance. They introduce a set of problems for which semantic annotation systems are usually employed (e.g. Annotate to Wikipedia (A2W) and Disambiguate to Wikipedia (D2W)) and provide metrics to evaluate systems in these contexts. However, these results might not be sufficient to distinguish the top performing annotators without a deeper statistical analysis, as we propose in this paper.

One of the main limitations of existing literature in semantic annotation evaluation \citep{www2013,weblog} is that it does not take into account the fact that the performance of a system may vary according to a specific task. By contrast, our evaluation aims at providing  experimental results on the performance of current semantic annotation systems for the three tasks (SA, NE, KW) with the objective of identifying annotators that are best suited for each of them. To achieve this, we rely on standard datasets that are experimentally selected for each task. We identify the statistical significance of systems' results using ANOVA (macro-average) and logistic regression (micro-average). 

\section{Research methodology}
\label{section:methodology}

\noindent In this paper, we address the following  research question:

\begin{quote}
\textit{RQ: How do linked data  annotators perform on the three tasks (SA, NE and KW)?}
\end{quote}

To answer this research question, we examine the overall spotting and disambiguation performance of linked data annotators in terms of macro and micro evaluation metrics. 

\subsection{Datasets}
\label{section:datasets}

We selected three different groups of datasets in English (a \textit{dataset} is a corpus where  mentions are  spotted and disambiguated according to  a gold standard).  Each dataset is focused on at least one  of three  tasks\footnote{The datasets and gold standards used in our evaluation are available \url{http://www.labowest.ca/AnnotatorsEvaluation}}  and uses DBpedia/Wikipedia as a background knowledge base.
These datasets include 1) AI and IITB for the  semantic annotation task, 2) MSNBC, for the evaluation of named entity annotation, and 3) the SemEval and Inspec datasets, which are used for the evaluation of keyword extraction.

\subsubsection{AI}
The AI\ corpus is a small set of documents composed of Wikipedia articles related to the artificial intelligence domain, which was used in previous experiments for ontology learning from text [20]. The gold standard was created by running all selected semantic annotators and evaluating the returned annotations as correct or incorrect. This evaluation was performed by two authors of this paper (two postdoctoral researchers).

\subsubsection{IITB}
IITB is a dataset proposed by \cite{kulkarni2009} which includes more than a hundred documents comprehensively annotated by human experts. Documents were collected from  popular Websites on sport, entertainment, health and science. In the literature, IITB is often used for the evaluation of named entity annotation, but in our evaluation we associate it to the  semantic annotation task, since it contains  annotations that go beyond named entities (eg: \textit{sniper}, \textit{militant}, \textit{October 7}, \textit{president of Afghanistan})

\subsubsection{MSNBC}

MSNBC is a small collection of news documents (18 documents) on different popular subjects such as sport, politics and technologies and was proposed by \cite{Cucerzan2007}. MSNBC is mainly focused on important named entities. However, an initial analysis revealed significant problems in the dataset, such as entities that are indicated in the gold standard, but not found in the documents,  entities cited in the documents, but absent from the gold standard, and, less frequently, incorrect entities specified in the gold standard. For the purpose of this research, we completely re-annotated the documents of this corpus, to obtained a gold standard more accurate than the original one.

\subsubsection{SemEval}
The SemEval dataset \citep{kim2010semeval}  is a standard benchmark for keyword extraction that associates key-phrases to  documents.  It  contains  244  scientific articles, usually composed of 6 to 8 pages. The articles cover different research areas of the ACM classification: Distributed Systems, Information Search and Retrieval, Distributed Artificial Intelligence, Multiagent Systems, Social and Behavioral Sciences and Economics. Most articles essentially cover the Computer Science domain (75\% of the documents) and the other documents  cover  the  Economy domain.
The gold standard includes key-phrases assigned by annotators (75\%) as well as key phrases assigned by the papers’ authors (25\%). The SemEval corpus is divided into a training dataset (144 articles, 2070 key-phrases) and testing dataset (100 articles, 1443 key-phrases). In our experiment, we consider the 244 articles as a single corpus.

\subsubsection{Inspec}

Inspec is a set of 2 000 documents and consists of abstracts from scientific journal papers. Each abstract has two sets of keywords assigned by a professional indexer. One is a set of controlled terms from the Inspec thesaurus, and the other one is an uncontrolled set of keywords that contain any suitable term identified by the indexer in texts. Both sets can contain keywords that are not found in the abstracts. In our evaluation,  we used only the uncontrolled set of keywords.

Table \ref{tab::stat:datasets} shows some descriptive statistics about the number of documents and the number of mentions in each dataset. We can notice that the Inspec dataset contains the highest number of mentions (it also contains many more documents), but the fewest number of mentions per document (due to the smaller size of documents and due to the fact that the keyword extraction task  identifies the most relevant keywords only). In the IITB and AI datasets, the average number of mentions per document is much higher than in the three other datasets. This is expected for the task of  semantic annotation. Finally, the average number of words in SemEval is much higher than in other datasets, but this value is somehow misleading, since these documents contain a high number of tokens that are not words (for example, elements of mathematical formulas).

\begin{table}[htbp]
\centering
\caption{Statistics on datasets. }
\begin{scriptsize}
\begin{tabular}{p{2cm}p{2cm}p{2cm}p{2cm}p{2cm}p{.9cm}}
\toprule
{\bf Corpus} 		&  \# doc & \# words/doc	&  \# mentions	&  \# mentions/doc &   Task \\
\toprule
AI 	        &   8 		& 1322 		& 713 		& 89.1  	   &  	SA\\
IITB 	    &   104 	& 640 		& 6866 		& 66.0  	   & 	SA \\
MSNBC 	    &   18 		& 544		& 392 		& 21.8  	   &  	NE \\
SemEval		&   244 	& 8022		& 3689 		& 15.1     &   KW\\
Inspec 	    &   2000  	& 124 	    & 19244 	& 9.6      &   KW\\
\bottomrule
\end{tabular}%
\end{scriptsize}
\label{tab::stat:datasets}%
\end{table}%

\subsection{Semantic annotators}
\label{section:systems}
In this section, we briefly present the semantic annotators selected for this study. For the purpose of our evaluation, we selected academically or industrially prominent semantic annotators available through a Web API.  

Table \ref{tab::systems} shows all the evaluated annotators categorized according to their best-suited task based on their description:  semantic annotation (SA), named entity annotation (NE) or keyword annotation (KW). We also indicate if the service is commercial, if the result of the annotation process may contain external entities that are not found in the text and, finally, the knowledge base that is used to disambiguate the entities. 

\begin{table*}[htbp]
  \small
	\centering
	\caption{Systems used in the current study}
	\begin{scriptsize}
	\begin{tabular}{lp{1cm}p{1.5cm}p{1cm}p{6cm}}
		\toprule
System & Cat. & Commerc. & External ent. & KB  \\
\hline		
Watson/Concepts   & SA & $\checkmark$ & $\checkmark$ & DBpedia  \\
Aylien/Concepts   & SA & $\checkmark$ &              & DBpedia \\
Babelfy/Concepts  & SA &              &              & DBpedia, Babelnet \\
Dandelion         & SA & $\checkmark$ &              & Wikipedia, \\
Spotlight         & SA &              &              & DBpedia \\
Open Calais       & SA & $\checkmark$ &              & Proprietary  \\
Tagme   		  & SA &              &              & Wikipedia  \\
Umbel   	      & SA &              &              & Umbel \\
Yahoo    	      & SA & $\checkmark$ &              & Wikipedia \\
Ambiverse         & SA & $\checkmark$ &              & Wikipedia \\
\hline

Aylien/Entities   & NE & $\checkmark$ &              & - \\
Babelfy/NE        & NE &              &              & DBpedia, Babelnet\\
Enrycher/NE       & NE &              &              & DBpedia, YAGO, OpenCyc \\
MeaningCloud/NE   & NE & $\checkmark$ &              & DBpedia \\
TextRazor         & NE & $\checkmark$ &              & Wikipdia, Freebase,  \\
Watson/NE         & NE & $\checkmark$ &              & DBpedia  \\

\hline
Aylien/KW         & KW & $\checkmark$ &              & - \\
Enrycher/KW       & KW &              & $\checkmark$ & - \\
MeaningCloud/KW   & KW & $\checkmark$ &              & - \\
Watson/KW         & KW & $\checkmark$ &              & - \\
\bottomrule
	\end{tabular}%
	\end{scriptsize}
	\label{tab::systems}%
\end{table*}%

Herafter, we describe the chosen semantic annotators. In some cases, the description is very brief due to the lack of published research on the semantic annotator.

Watson\footnote{https://www.ibm.com/watson/services/natural-language-understanding/} APIs employ a set of deep linguistic parsing methods and statistical language processing techniques for performing semantic annotation. Various APIs are available, among which three are relevant to our research objectives: named entity extraction (Watson/NE), keyword extraction (Watson/KW), and concept extraction (Watson/Concepts). The named entity extractor  (Watson/NE) is able to disambiguate the detected entities and resolve co-references. Entities are linked to various datasets on the Linked Open Data Cloud (LOD). Keyword extraction (Watson/KW) produces a list of key-phrases without any linkage to an external knowledge base (i.e. without disambiguation). Concept extraction (Watson/Concepts) produces a list of concepts, that is, topics that are not necessarily mentioned in the text, along with their corresponding links on the LOD.

Aylien\footnote{http://docs.aylien.com/} is another commercial product that offers two services that are relevant for our study. One is the concept extraction service and the other is the entity extraction service, which not only extracts named entities, but also keywords. Since these results correspond to different tasks in our framework, we analyzed them separately. Note that this second service does not provide any disambiguation for the annotated entities.

On its website\footnote{http://babelfy.org/}, Babelfy \citep{moro2014} is defined as "a unified, multilingual, graph-based approach to Entity Linking and Word Sense Disambiguation based on a loose identification of candidate meanings coupled with a densest subgraph heuristic which selects high-coherence semantic interpretations". Babelfy is based on BabelNet, a multilingual semantic network. 

Dandelion\footnote{https://dandelion.eu/}  offers several text analysis services for many languages: entity extraction, text similarity, text classification, language detection and sentiment analysis. Only the first one is of interest for our study. 

Open Calais\footnote{http://www.opencalais.com/} is a service offered by Thomson Reuters. It can detect different kinds of entities, which are disambiguated with a proprietary knowledge base. It can also detect events, relationships and topics.

DBpedia Spotlight\footnote{https://github.com/dbpedia/dbpedia/wiki} \citep{Mendes2011} is a configurable annotator that is linked to DBpedia (we used the default values).  After the spotting phase, DBpedia Spotlight  pre-ranks DBpedia concept candidates for each spotted key-phrase in text. It uses a similarity score to determine which candidate concept is the most relevant. The similarity score takes into account the context of the phrase (a window of words around the phrase) and the context of each candidate concept.

Tagme\footnote{http://tagme.di.unipi.it/} is a semantic annotator mainly designed for analyzing short texts such as tweets \citep{ferragina2010tagme}, but it has also been reported to perform well on longer documents \citep{cornolti2013framework}. Tagme tokenizes a given text and finds candidate spots from token sequences of up to six words. It uses a set of heuristics and probability and coherence measures to decide which spotted candidates should be considered for disambiguation and which spot must be pruned from the result set. Tagme returns all annotations in a text plus their corresponding relevance scores according to the text topic. 

Umbel\footnote{http://www.umbel.org/web-services/tagger-concept-noun/} offers two tagging services. One tries to detect concepts from the Umbel ontology in texts, and the other is restricted to noun phrases. It is the latter that has been used in this study. Note that by default, Umbel does not apply any stemming.

Yahoo Content Analysis API\footnote{https://developer.yahoo.com/contentanalysis/} annotates entities and concepts and also provides a ranking of these entities and concepts, according to their overall relevance. Access to the service is achieved through the Yahoo Query Language (YQL), a SQL-like language that enables querying, filtering, and combining data across the web.

TextRazor\footnote{https://www.textrazor.com/}  offers many services for the extraction of information from text. It also enables customization by using Prolog rules. In our study, we only use the entity recognition service.

Enrycher\footnote{http://ailab.ijs.si/tools/enrycher/} provides deep and shallow text processing services. We used the two following services: named entity resolution and keyword detection.

Finally, MeaningCloud\footnote{http://www.meaningcloud.com/} offers several text analysis services for many languages: topic extraction, text classification, sentiment analysis and text clustering. The topic extraction service detects and disambiguates named entities, and it can also extracts keywords. MeaningCloud offers the possibility of adding your own dictionaries in its annotation services.

\subsection{Evaluation metrics}
\label{section:metrics} 

The following metrics are used for the evaluation of the semantic annotators:
\begin{itemize}
  \item \textit{Precision}: the ratio of the number of correct items returned by the annotator over the total number of items returned by the annotator.
  \item \textit{Recall}: the ratio of the number of correct items returned by the annotator over the total number of  items specified in the gold standard. 

  \item \textit{F-measure}: harmonic mean of precision and recall, $2\times\frac{P \times R}{P + R}$.
\end{itemize}

There are two ways of averaging the performance values: macro-average and micro-average. The first one consists in computing the metrics for each document and then averaging  these values. In the second one, we take all mentions, without considering their source document, and compute the ratio of hits among these mentions. In our case, both approaches are relevant. The recognition of an entity in a document may depend on the overall context of the document\footnote{In fact, it is the case for many systems in our evaluation}. For example, the occurrence of \textit{Washington} in a document may refer to many different entities, and the document helps disambiguate its meaning. In this case, where the document represents the relevant context, macro-average should be used. Nevertheless, in many systems, a limited context of few words around the mention is used. In this case, each mention may be taken individually, and micro-average is well-suited for the evaluation. An important advantage of the micro-average is that it is less sensitive to the number of documents and their length. In this paper, we evaluate the systems with both approaches. 

\begin{table}[htbp]
\scriptsize
\centering
\caption{Number of annotations extracted by each system, for each corpus}
\begin{tiny}
\begin{tabular}{p{1.2cm}llp{0.75cm}p{0.75cm}ll}
\toprule
System 	& AI 	& IITB 	& MSNBC 	& SemEval 	& Inspec 	& Average  \\
\toprule
Ambiverse    & 76     & 1716   & 173    & 12433  & 2791   & 3438   \\
AylienCon    & 260    & 2022   & 152    & 11860  & 5241   & 3907   \\
AylienEnt    & 68     & 1332   & 147    & 12283  & 2726   & 3311   \\
AylienKW     & 143    & 2049   & 198    & 4607   & 37886  & 8977   \\
BabelCon     & 1346   & 11853  & 932    & 44597  & 58843  & 23514  \\
BabelNE      & 113    & 2395   & 213    & 5132   & 3398   & 2250   \\
Dandelion    & 1135   & 5464   & 402    & 62580  & 33359  & 20588  \\
EnrychKW     & 51     & 663    & 71     & 1119   & 14445  & 3270   \\
EnrychNE     & 54     & 1031   & 111    & 2996   & 2253   & 1289   \\
MCloudEnt    & 115    & 1988   & 169    & 16547  & 3820   & 4528   \\
MCloudKW     & 656    & 4983   & 374    & 38754  & 25605  & 14074  \\
OpCalais     & 159    & 2286   & 222    & 16952  & 5345   & 4993   \\
Spotlight    & 377    & 2933   & 210    & 22821  & 10040  & 7276   \\
Tagme        & 1974   & 15212  & 1079   & 120291 & 70616  & 41834  \\
TextRazor    & 989    & 6284   & 624    & 67826  & 6498   & 16444  \\
Umbel        & 444    & 3184   & 257    & 26529  & 17803  & 9643   \\
WatsonCon    & 63     & 809    & 79     & 1835   & 12917  & 3141   \\
WatsonKW     & 333    & 4710   & 393    & 10592  & 40346  & 11275  \\
WatsonNE     & 84     & 2191   & 203    & 9507   & 3198   & 3037   \\
Yahoo        & 68     & 932    & 95     & 2040   & 12979  & 3223   \\
\hline
Average      & 425    & 3702   & 305    & 24565  & 18505  \\
\bottomrule

\end{tabular}
\end{tiny}
\label{tab::stat:extracted_kw}
\end{table}

We ran all semantic annotators on all the available datasets. Table \ref{tab::stat:extracted_kw} shows the total number of annotations extracted by each semantic annotator. We can notice considerable variations across systems and across datasets. For example, Tagme returns 5.7 times more annotations than DBpedia Spotlight, on average (41834 vs 7276). Similarly, Babelfy/Concepts, Dandelion, MeaningCloud/KW, Tagme, TextRazor and Watson/KW return more than 10 000 annotations on the average, while other systems return a much lower number of annotations (1289 for Enrycher/NE, and 2250 for Babelfy/NE). We may expect here  that systems with the largest number of annotations will exhibit a high recall. We can also note that much more annotations are extracted from SemEval compared to other datasets. In fact, in this corpus, there are much more extracted keywords on the average than the number of correct mentions in the gold standard (more than 20000 on average, compared to 3689 in the gold standard). We thus expect a very low precision for this corpus. AI and MSNBC are the smallest datasets in terms of numbers of annotations, which is expected, since the number of correct mentions in their gold standards is also very low compared to other datasets.

\section{Evaluation using macro-average}
\label{section:evaluation:macro}

In this section, we use the macro-average to evaluate the performance of the systems for the two main steps of semantic annotation, namely the spotting step and the disambiguation step. In each step, we analyze the precision and recall of semantic annotators for  the SA and NE tasks. In the  KW task, there is not any disambiguation step, so only spotting is evaluated. All mentions returned by annotators and all the ones provided in gold standards are stemmed using an implementation of the Porter stemmer. As an example, two key-phrases ’parallel processes’ and ’parallel processing’ are matched to the gold standard entry "parallel process" because all have the same stem. This approach seems reasonable for the spotting phase, as we consider all these alternatives as valid  mentions.  If an entity is spotted more than once in a text, it appears only  once in the gold standard. This means that we evaluate the capability of spotting at least one occurrence of each relevant mention. 

\begin{table*}
\label{precision:spotting:macro}
\centering
\caption{Ranked precision values (macro-averages) for the spotting step. Dashlines  indicate the borders  of groups of systems that are indistinguishable, according to ANOVA statistical analysis with Tukey HSD post hoc test.}
\scriptsize
\noindent
\begin{tabular}{lllll}
\begin{tabular}{ll}
\toprule
\multicolumn{2}{c}{AI} \\
\hline
WatsonCon & 0.92  \\
Yahoo & 0.80   \\
AylienCon & 0.73  \\
Ambiverse & 0.73  \\
Spotlight & 0.68  \\
OpCalais & 0.65  \\
MCloudEnt & 0.63  \\
AylienEnt & 0.58  \\
\hdashline[1pt/1pt]
WatsonKW & 0.54  \\
WatsonNE & 0.53  \\
TextRazor & 0.49  \\
BabelNE & 0.48  \\
EnrychNE & 0.46  \\
Dandelion & 0.43   \\
AylienKW & 0.43   \\
BabelCon & 0.32   \\
Tagme & 0.26  \\
MCloudKW & 0.25   \\
EnrychKW & 0.25   \\
Umbel & 0.24   \\
\bottomrule
\end{tabular}
 & 
\begin{tabular}{ll}
\toprule
\multicolumn{2}{c}{IITB} \\
\hline
AylienCon & 0.74 \\
Spotlight & 0.69 \\
\hdashline[1pt/1pt]
Dandelion & 0.63 \\
\hdashline[1pt/1pt]
EnrychNE & 0.54  \\
Yahoo & 0.52  \\
TextRazor & 0.50  \\
Ambiverse & 0.48  \\
OpCalais & 0.47  \\
WatsonNE & 0.47  \\
WatsonCon & 0.45  \\
\hdashline[1pt/1pt]
AylienEnt & 0.44 \\
AylienKW & 0.41  \\
MCloudEnt & 0.41  \\
BabelNE & 0.41  \\
MCloudKW & 0.37  \\
BabelCon & 0.37  \\
\hdashline[1pt/1pt]

Umbel & 0.34  \\
Tagme & 0.33   \\
\hdashline[1pt/1pt]
WatsonKW & 0.22   \\
EnrychKW & 0.17   \\
\bottomrule
\end{tabular}
 & 
\begin{tabular}{ll}
\toprule
\multicolumn{2}{c}{MSNBC} \\
\hline
Ambiverse & 0.74  \\
EnrychNE & 0.70  \\
AylienEnt & 0.64  \\
AylienCon & 0.58  \\
\hdashline[1pt/1pt]
MCloudEnt & 0.54  \\
WatsonNE & 0.54  \\
Spotlight & 0.47  \\
BabelNE & 0.47  \\
OpCalais & 0.41  \\
Yahoo & 0.38  \\
\hdashline[1pt/1pt]
Dandelion & 0.33  \\
WatsonCon & 0.29  \\
TextRazor & 0.25  \\
\hdashline[1pt/1pt]
WatsonKW & 0.15  \\
AylienKW & 0.13  \\
Tagme & 0.089  \\
BabelCon & 0.054  \\
MCloudKW & 0.014  \\
EnrychKW & 0.0078  \\
Umbel & 0.0044  \\

\bottomrule
\end{tabular}
 & 
\begin{tabular}{ll}
\toprule
\multicolumn{2}{c}{SemEval} \\
\hline
Yahoo & 0.27  \\
\hdashline[1pt/1pt]
WatsonCon & 0.14  \\
\hdashline[1pt/1pt]
WatsonKW & 0.11  \\
\hdashline[1pt/1pt]
AylienCon & 0.045  \\
Spotlight & 0.032  \\
\hdashline[1pt/1pt]
BabelCon & 0.027  \\
OpCalais & 0.025  \\
Dandelion & 0.021  \\
AylienKW & 0.02  \\
WatsonNE & 0.017  \\
TextRazor & 0.017  \\
Ambiverse & 0.014  \\
MCloudEnt & 0.013  \\
AylienEnt & 0.012  \\
Tagme & 0.010  \\
\hdashline[1pt/1pt]
EnrychKW & 0.0086  \\
Umbel & 0.0086  \\
BabelNE & 0.0085  \\
MCloudKW & 0.0084  \\
EnrychNE & 0.0059  \\
\bottomrule
\end{tabular}
 & 
\begin{tabular}{ll}
\toprule
\multicolumn{2}{c}{Inspec} \\
\hline
Yahoo & 0.33  \\
\hdashline[1pt/1pt]
AylienCon & 0.30  \\
\hdashline[1pt/1pt]
OpCalais & 0.27  \\
Spotlight & 0.26  \\
\hdashline[1pt/1pt]
WatsonKW & 0.23  \\
\hdashline[1pt/1pt]
Dandelion & 0.15  \\
\hdashline[1pt/1pt]
MCloudEnt & 0.12  \\
\hdashline[1pt/1pt]
WatsonNE & 0.092  \\
BabelCon & 0.088  \\
BabelNE & 0.086  \\
Ambiverse & 0.085  \\
Tagme & 0.084  \\
AylienEnt & 0.074  \\
\hdashline[1pt/1pt]
WatsonCon & 0.065  \\
AylienKW & 0.057  \\
EnrychNE & 0.048  \\
MCloudKW & 0.044  \\
\hdashline[1pt/1pt]
TextRazor & 0.036  \\
Umbel & 0.022  \\
\hdashline[1pt/1pt]
EnrychKW & 0.011  \\
\bottomrule
\end{tabular}
\\
3\end{tabular}
\end{table*}

\subsection{The spotting step}

Table 4 presents the precision values obtained for all datasets, considering all spotted entities returned by systems (without any filtering). We can see that the AI corpus cannot really be used to distinguish systems' results as there are only two performance groups based on ANOVA. This is probably due to the small number of documents (8 documents only). With the other corpora, the situation is clearer: Aylien/Concepts and DBpedia Spotlight are the most precise APIs on the IITB corpus, which corresponds to the NE task. On SemEval and Inspec, results are quite low, with Yahoo being the top performer in both cases. Some systems are among the best ones on almost all corpora: Yahoo, Aylien/Concepts and DBpedia Spotlight. We can also note that, as expected, the precision observed on SemEval and Inspec is very low for all systems (remember that the SemEval corpus is the one with the largest number of annotations returned by the systems, thus increasing the probability of an incorrect annotation).  
Overall, the results in Table 4 show us, for instance, that a precision of 0.54 is in the same group as a precision of 0.45 on the IITB corpus. A shallower observation of those results might have led to a false conclusion, i.e. that EnrychNE outperforms by far WatsonCon, which is not the case.

\begin{table*}[htbp]
\label{recall:spotting:macro}
\centering
\caption{Ranked recall values (macro-averages) for the spotting step. Dashlines  indicate the borders of
groups of systems that are indistinguishable, according to ANOVA statistical analysis with Tukey HSD post hoc test.}
\scriptsize
\noindent
\begin{tabular}{lllll}
\begin{tabular}{ll}
\toprule
\multicolumn{2}{c}{AI} \\
\hline
Dandelion & 0.6  \\
TextRazor & 0.59  \\
Tagme & 0.59  \\
BabelCon & 0.55  \\
\hdashline[1pt/1pt]
WatsonKW & 0.32  \\
Spotlight & 0.31  \\
AylienCon & 0.24  \\
MCloudKW & 0.19  \\
\hdashline[1pt/1pt]
Umbel & 0.12  \\
AylienKW & 0.12  \\
OpCalais & 0.12  \\
WatsonCon & 0.11  \\
Yahoo & 0.11  \\
BabelNE & 0.072  \\
MCloudEnt & 0.07  \\
WatsonNE & 0.062  \\
Ambiverse & 0.048  \\
AylienEnt & 0.041  \\
EnrychNE & 0.026  \\
EnrychKW & 0.025  \\
\bottomrule
\end{tabular}
 & 
\begin{tabular}{ll}
\toprule
\multicolumn{2}{c}{IITB} \\
\hline
Tagme & 0.72  \\
\hdashline[1pt/1pt]
BabelCon & 0.63  \\
\hdashline[1pt/1pt]
Dandelion & 0.48  \\
TextRazor & 0.47  \\
\hdashline[1pt/1pt]
Spotlight & 0.3  \\
MCloudKW & 0.27  \\
\hdashline[1pt/1pt]
AylienCon & 0.23  \\
WatsonKW & 0.18  \\
\hdashline[1pt/1pt]
BabelNE & 0.18  \\
OpCalais & 0.17  \\
WatsonNE & 0.16  \\
Umbel & 0.16  \\
AylienKW & 0.15  \\
Ambiverse & 0.15  \\
MCloudEnt & 0.14  \\
\hdashline[1pt/1pt]
EnrychNE & 0.1  \\
AylienEnt & 0.1  \\
Yahoo & 0.089  \\
WatsonCon & 0.062  \\
\hdashline[1pt/1pt]
EnrychKW & 0.015  \\
\bottomrule
\end{tabular}
 & 
\begin{tabular}{ll}
\toprule
\multicolumn{2}{c}{MSNBC} \\
\hline
TextRazor & 0.81  \\
Ambiverse & 0.79  \\
Dandelion & 0.71  \\
BabelNE & 0.67  \\
WatsonNE & 0.66  \\
Tagme & 0.64  \\
\hdashline[1pt/1pt]
Spotlight & 0.62  \\
MCloudEnt & 0.57  \\
AylienEnt & 0.57  \\
AylienCon & 0.56  \\
OpCalais & 0.54  \\
EnrychNE & 0.48  \\
\hdashline[1pt/1pt]
WatsonKW & 0.39  \\
BabelCon & 0.33  \\
Yahoo & 0.24  \\
\hdashline[1pt/1pt]
AylienKW & 0.18  \\
WatsonCon & 0.17  \\
MCloudKW & 0.027  \\
Umbel & 0.014  \\
EnrychKW & 0.0061  \\
\bottomrule
\end{tabular}
 & 
\begin{tabular}{ll}
\toprule
\multicolumn{2}{c}{SemEval} \\
\hline
Dandelion & 0.34  \\
BabelCon & 0.33  \\
WatsonKW & 0.31  \\
\hdashline[1pt/1pt]
Tagme & 0.29  \\
TextRazor & 0.28  \\
\hdashline[1pt/1pt]
Spotlight & 0.2  \\
\hdashline[1pt/1pt]
Yahoo & 0.15  \\
AylienCon & 0.15  \\
\hdashline[1pt/1pt]
OpCalais & 0.12  \\
MCloudKW & 0.09  \\
\hdashline[1pt/1pt]
WatsonCon & 0.077  \\
MCloudEnt & 0.068  \\
Umbel & 0.064  \\
WatsonNE & 0.045  \\
AylienEnt & 0.044  \\
\hdashline[1pt/1pt]
AylienKW & 0.024  \\
Ambiverse & 0.024  \\
BabelNE & 0.012  \\
EnrychNE & 0.0081  \\
EnrychKW & 0.0036  \\
\bottomrule
\end{tabular}
 & 
\begin{tabular}{ll}
\toprule
\multicolumn{2}{c}{Inspec} \\
\hline
WatsonKW & 0.47  \\
\hdashline[1pt/1pt]
Tagme & 0.3  \\
\hdashline[1pt/1pt]
BabelCon & 0.27  \\
\hdashline[1pt/1pt]
Yahoo & 0.25  \\
Dandelion & 0.24  \\
\hdashline[1pt/1pt]
Spotlight & 0.14  \\
AylienKW & 0.13  \\
\hdashline[1pt/1pt]
AylienCon & 0.093  \\
OpCalais & 0.093  \\
\hdashline[1pt/1pt]
MCloudKW & 0.063  \\
WatsonCon & 0.053  \\
\hdashline[1pt/1pt]
TextRazor & 0.04  \\
MCloudEnt & 0.035  \\
BabelNE & 0.026  \\
\hdashline[1pt/1pt]
WatsonNE & 0.024  \\
Ambiverse & 0.021  \\
Umbel & 0.021  \\
AylienEnt & 0.019  \\
EnrychNE & 0.01  \\
EnrychKW & 0.0097  \\
\bottomrule
\end{tabular}
\\
\end{tabular}
\end{table*}

Table 5 presents the recall values. We can see that the best performing systems are not the same as in Table 4. Interestingly, Tagme and Babelfy/Concepts are among the worst systems in terms of precision, but among the best ones if we consider recall. This is expected, considering the high number of annotations returned by these system, which necessarily favours recall over precision.  Yahoo, which is the best API in both SemEval and Inspec in terms of precision, does not perform so well in terms of recall on these two same corpora. At the opposite, Watson/KW obtains low precision values on these corpora, but performs very well in term of recall. Note that a recall of 0.47 puts this system clearly in a dominant position compared to the other ones.

The trade-off between precision and recall can be observed in Table 6. For example, considering the IITB corpus, Dandelion, which is not among the top systems in precision and recall, is clearly the best one if we consider its F-score. Looking at the results for the two corpora related to the KW task, we see that Yahoo and Watson/KW are the top systems despite low F-scores. Interestingly, we can see that for both SA and KW tasks, the top-performing systems are almost the same for both corpora (TextRazor, Dandelion, Spotlight, Babel/Concepts,  Aylien/Concepts and Tagme for the SA task; Yahoo and Watson/KW for the KW task). Here we observe that a keyword extractor Watson/KW is among the second group of systems for the AI corpus, despite the fact that this corpus is related to the SA task.

\begin{table*}[htbp]
\label{zzz}
\centering
\caption{Ranked F-score values (macro-averages) for spotting step. Dashlines indicate the borders of groups of systems that are indistinguishable, according to ANOVA statistical analysis with Tukey HSD post hoc test.}
\scriptsize
\noindent
\begin{tabular}{lllll}
\begin{tabular}{ll}
\toprule
\multicolumn{2}{c}{AI} \\
\hline
TextRazor & 0.52  \\
Dandelion & 0.49  \\
Spotlight & 0.42  \\
BabelCon & 0.4  \\
\hdashline[1pt/1pt]
WatsonKW & 0.37  \\
AylienCon & 0.36  \\
Tagme & 0.35  \\
\hdashline[1pt/1pt]
MCloudKW & 0.22  \\
OpCalais & 0.2  \\
WatsonCon & 0.19  \\
Yahoo & 0.18  \\
AylienKW & 0.17  \\
Umbel & 0.16  \\
BabelNE & 0.12  \\
MCloudEnt & 0.12  \\
WatsonNE & 0.11  \\
Ambiverse & 0.084  \\
\hdashline[1pt/1pt]
AylienEnt & 0.074  \\
EnrychNE & 0.045  \\
EnrychKW & 0.044  \\
\bottomrule
\end{tabular}
 & 
\begin{tabular}{ll}
\toprule
\multicolumn{2}{c}{IITB} \\
\hline
Dandelion & 0.53  \\
\hdashline[1pt/1pt]
TextRazor & 0.46  \\
BabelCon & 0.45  \\
Tagme & 0.44  \\
Spotlight & 0.41  \\
\hdashline[1pt/1pt]
AylienCon & 0.34  \\
MCloudKW & 0.3  \\
\hdashline[1pt/1pt]
OpCalais & 0.23  \\
WatsonNE & 0.21  \\
BabelNE & 0.21  \\
AylienKW & 0.21  \\
Umbel & 0.2  \\
Ambiverse & 0.2  \\
WatsonKW & 0.18  \\
MCloudEnt & 0.18  \\
\hdashline[1pt/1pt]
EnrychNE & 0.15  \\
AylienEnt & 0.14  \\
Yahoo & 0.14  \\
WatsonCon & 0.1  \\
\hdashline[1pt/1pt]
EnrychKW & 0.021  \\
\bottomrule
\end{tabular}
 & 
\begin{tabular}{ll}
\toprule
\multicolumn{2}{c}{MSNBC} \\
\hline
Ambiverse & 0.75  \\
\hdashline[1pt/1pt]
WatsonNE & 0.57  \\
AylienEnt & 0.56  \\
AylienCon & 0.55  \\
EnrychNE & 0.54  \\
BabelNE & 0.53  \\
MCloudEnt & 0.53  \\
Spotlight & 0.52  \\
OpCalais & 0.45  \\
Dandelion & 0.44  \\
\hdashline[1pt/1pt]
TextRazor & 0.37  \\
Yahoo & 0.26  \\
\hdashline[1pt/1pt]
WatsonKW & 0.2  \\
WatsonCon & 0.2  \\
Tagme & 0.15  \\
AylienKW & 0.14  \\
BabelCon & 0.09  \\
\hdashline[1pt/1pt]
MCloudKW & 0.015  \\
Umbel & 0.0067  \\
EnrychKW & 0.0067  \\
\bottomrule
\end{tabular}
 & 
\begin{tabular}{ll}
\toprule
\multicolumn{2}{c}{SemEval} \\
\hline
Yahoo & 0.17  \\
\hdashline[1pt/1pt]
WatsonKW & 0.15  \\
\hdashline[1pt/1pt]
WatsonCon & 0.079  \\
\hdashline[1pt/1pt]
AylienCon & 0.059  \\
Spotlight & 0.05  \\
BabelCon & 0.048  \\
\hdashline[1pt/1pt]
Dandelion & 0.038  \\
OpCalais & 0.033  \\
TextRazor & 0.028  \\
\hdashline[1pt/1pt]
Tagme & 0.018  \\
WatsonNE & 0.016  \\
AylienKW & 0.014  \\
MCloudEnt & 0.014  \\
MCloudKW & 0.012  \\
AylienEnt & 0.011  \\
Umbel & 0.01  \\
Ambiverse & 0.0089  \\
BabelNE & 0.0053  \\
\hdashline[1pt/1pt]
EnrychNE & 0.0031  \\
EnrychKW & 0.0021  \\
\bottomrule
\end{tabular}
 & 
\begin{tabular}{ll}
\toprule
\multicolumn{2}{c}{Inspec} \\
\hline
WatsonKW & 0.3  \\
\hdashline[1pt/1pt]
Yahoo & 0.27  \\
\hdashline[1pt/1pt]
Dandelion & 0.17  \\
Spotlight & 0.17  \\
\hdashline[1pt/1pt]
AylienCon & 0.13  \\
BabelCon & 0.13  \\
Tagme & 0.13  \\
OpCalais & 0.12  \\
\hdashline[1pt/1pt]
AylienKW & 0.076  \\
\hdashline[1pt/1pt]
WatsonCon & 0.055  \\
MCloudEnt & 0.048  \\
MCloudKW & 0.048  \\
\hdashline[1pt/1pt]
BabelNE & 0.035  \\
WatsonNE & 0.033  \\
TextRazor & 0.033  \\
Ambiverse & 0.031  \\
AylienEnt & 0.027  \\
\hdashline[1pt/1pt]
Umbel & 0.02  \\
EnrychNE & 0.016  \\
EnrychKW & 0.0093  \\
\bottomrule
\end{tabular}
\\
\end{tabular}
\end{table*}%



Now considering MSNBC corpus, most of the best performing systems are NE systems, with the notable exception of Ambiverse, the top API, which is also the only one that appears among the best ones for both precision and recall. 

Considering the SA and NE tasks, we can conclude that globally, the results correspond to the expectations, that is, the best performing systems on each corpus are the ones whose task is related to this corpus. For the KW task, the situation is quite different. By looking at the two corpora associated to this task, we see that only Watson/KW appears among the best-performing systems, due mainly to its very good recall score. Keyword extraction seems to be a difficult task, with F-scores lower than the ones obtained in the other tasks. Interestingly, Yahoo, the best performing system on SemEval, and second-best on Inspec, is not specifically designed as a KW extractor.

\subsection{The disambiguation step}
%
%

In this section, we evaluate the disambiguation capability of semantic annotators. Three datasets are used for this task: AI, IITB and MSNBC. We also consider only full semantic annotators and named entity annotators, as keyword extractors usually do not return any disambiguation information with their key-phrases. Also we restrict the evaluation to the systems that output links to Wikipedia or DBpedia. An annotation is considered correct if the pair $\langle m, e\rangle$, where $m$ and $e$ are the textual mention and the linked entity, respectively, is found as such in the Gold Standard. If the mention $m$ is found in the gold standard but annotated with the wrong entity by the system, it is considered as a miss. If a mention returned by a system is not contained in the gold standard, it is ignored, since there is no way of determining whether the entity is correctly disambiguated. To check the validity of the disambiguated entity, we take into account Wikipedia redirect links. Thus, a system that would return \textit{London\_Heathrow\_Airport} will be considered correct even if the entity  in the gold standard is  \textit{Heathrow\_Airport}. 
 Put simply, we are computing the ratio of correctly disambiguated entities among the correctly spotted mentions, that is, we evaluate the probability P(Disambiguating | Spotting). Practically, this means that, for each annotator, mentions that are not correctly spotted by the annotator are removed from the gold standard, leading to a reduced gold standard.  Here recall is equal to precision, since all tested systems provide at least one entity for each spotted mention and these mentions are the same in the reduced gold standard.
 

\begin{table}[htbp]
\label{disambiguation}
\centering
\footnotesize
\caption{Precision obtained for the disambiguation task. Superscript letters group system whose performances are statistically indistinguishable, using ANOVA with post hoc Tukey analysis.}
\scriptsize
\begin{tabular}{p{2cm}p{1cm}p{1cm}p{1cm}p{1cm}}
\multicolumn{4}{c}{Precision} \\
\toprule
              & AI         & IITB       & MSNBC      \\
\hline
WatsonCon     & 1.00$^a$   & 0.94$^a$   & 0.96$^a$   \\
EnrychNE      & 1.00$^a$   & 0.80$^b$   & 0.75$^b$   \\
BabelNE       & 0.94$^a$   & 0.74$^b$   & 0.82$^a$   \\
TextRazor     & 0.92$^a$   & 0.75$^b$   & 0.79$^a$   \\
Tagme         & 0.91$^a$   & 0.60$^d$   & 0.87$^a$   \\
Yahoo         & 0.91$^a$   & 0.74$^b$   & 0.75$^b$   \\
AylienCon     & 0.88$^a$   & 0.76$^b$   & 0.86$^a$   \\
Dandelion     & 0.87$^a$   & 0.70$^c$   & 0.89$^a$   \\
Ambiverse     & 0.86$^a$   & 0.77$^b$   & 0.78$^a$   \\
Spotlight     & 0.86$^a$   & 0.74$^b$   & 0.85$^a$   \\
BabelCon      & 0.73$^b$   & 0.52$^d$   & 0.16$^c$   \\
WatsonNE      & 0.37$^c$   & 0.55$^d$   & 0.60$^b$   \\
\bottomrule
\end{tabular}
\vspace{0.25cm}
\end{table}%
\begin{table}[htbp]
\label{disambiguation+spotting}
\centering
\footnotesize
\caption{Results obtained for the full annotation process (spotting + disambiguation). Superscript letters group systems whose performances are statistically distinguishable, using ANOVA with post hoc Tukey analysis.}
\scriptsize
\begin{tabular}{p{2cm}p{1cm}p{1cm}p{1cm}p{1cm}}
\multicolumn{4}{c}{Precision} \\
\toprule
              & AI         & IITB       & MSNBC      \\
\hline
WatsonCon     & 0.85$^a$   & 0.23$^c$   & 0.26$^b$   \\
EnrychNE      & 0.59$^a$   & 0.45$^b$   & 0.53$^a$   \\
AylienCon     & 0.55$^b$   & 0.56$^a$   & 0.50$^a$   \\
Spotlight     & 0.49$^a$   & 0.50$^a$   & 0.39$^b$   \\
BabelNE       & 0.48$^b$   & 0.30$^c$   & 0.38$^b$   \\
Ambiverse     & 0.45$^b$   & 0.39$^b$   & 0.58$^a$   \\
Yahoo         & 0.41$^b$   & 0.39$^b$   & 0.29$^b$   \\
TextRazor     & 0.36$^b$   & 0.37$^b$   & 0.20$^c$   \\
Dandelion     & 0.31$^b$   & 0.43$^b$   & 0.28$^b$   \\
Tagme         & 0.22$^c$   & 0.19$^d$   & 0.07$^d$   \\
WatsonNE      & 0.22$^c$   & 0.27$^c$   & 0.33$^b$   \\
BabelCon      & 0.18$^c$   & 0.19$^d$   & 0.01$^d$   \\
\bottomrule
\end{tabular}
\vspace{0.25cm}

\begin{tabular}{p{2cm}p{1cm}p{1cm}p{1cm}p{1cm}}
\multicolumn{4}{c}{Recall} \\
\toprule
              & AI         & IITB       & MSNBC      \\
\hline
Tagme         & 0.59$^a$   & 0.42$^a$   & 0.54$^a$   \\
Dandelion     & 0.56$^a$   & 0.34$^b$   & 0.62$^a$   \\
TextRazor     & 0.55$^a$   & 0.38$^a$   & 0.65$^a$   \\
BabelCon      & 0.39$^b$   & 0.33$^b$   & 0.05$^c$   \\
Spotlight     & 0.27$^b$   & 0.21$^c$   & 0.52$^a$   \\
AylienCon     & 0.21$^c$   & 0.16$^d$   & 0.48$^a$   \\
WatsonCon     & 0.12$^c$   & 0.03$^f$   & 0.13$^c$   \\
Yahoo         & 0.07$^d$   & 0.07$^e$   & 0.17$^c$   \\
BabelNE       & 0.06$^d$   & 0.13$^d$   & 0.54$^a$   \\
Ambiverse     & 0.04$^d$   & 0.13$^d$   & 0.61$^a$   \\
EnrychNE      & 0.04$^d$   & 0.08$^e$   & 0.36$^b$   \\
WatsonNE      & 0.03$^d$   & 0.09$^e$   & 0.38$^b$   \\
\bottomrule
\end{tabular}
\vspace{0.25cm}

\begin{tabular}{p{2cm}p{1cm}p{1cm}p{1cm}p{1cm}}
\multicolumn{4}{c}{F-score} \\
\toprule
              & AI         & IITB       & MSNBC      \\
\hline
TextRazor     & 0.43$^a$   & 0.37$^a$   & 0.29$^b$   \\
Dandelion     & 0.39$^a$   & 0.37$^a$   & 0.37$^b$   \\
Spotlight     & 0.33$^a$   & 0.28$^b$   & 0.43$^b$   \\
Tagme         & 0.31$^a$   & 0.26$^b$   & 0.12$^c$   \\
AylienCon     & 0.29$^b$   & 0.24$^c$   & 0.47$^a$   \\
BabelCon      & 0.24$^b$   & 0.23$^c$   & 0.01$^d$   \\
WatsonCon     & 0.20$^b$   & 0.05$^f$   & 0.13$^c$   \\
Yahoo         & 0.12$^c$   & 0.10$^e$   & 0.18$^c$   \\
BabelNE       & 0.10$^c$   & 0.16$^d$   & 0.43$^b$   \\
EnrychNE      & 0.07$^c$   & 0.12$^d$   & 0.40$^b$   \\
Ambiverse     & 0.06$^c$   & 0.16$^d$   & 0.57$^a$   \\
WatsonNE      & 0.03$^c$   & 0.11$^e$   & 0.33$^b$   \\
\bottomrule
\end{tabular}%
\end{table}%

Table 7 shows the scores obtained for disambiguation where we obtained very few groups using ANOVA with post hoc Tukey analysis (3 for AI, 4 for IITB and 3 for MSNBC).
We can note that Watson/Concept outperforms all other systems on IITB, and is among the leading systems for the two other corpora. We can also note that the systems are usually rather good at disambiguating an entity once the spot has been found. Interestingly, we observe that Tagme is in the first group for AI and MSNBC but in the last one for IITB.

Let us now consider the full annotation process, that is, the combination of spotting and disambiguation. In this case, an annotation is considered as a hit if it has been both correctly spotted and correctly disambiguated, and a miss if has not been correctly spotted (and should be according to gold standard) or if it has been correctly spotted but incorrectly disambiguated. We see that the results are not as good (see Table 8). For example precision values are much lower, especially on the IITB and MSNBC corpora. Watson/Concepts, which performs well on all corpora for disambiguation, outperforms the other systems only on AI for the full annotation process. In fact, we see that there is no system that really outperforms the other ones in terms of precision. 
In terms of recall, Tagme and TextRazor are among the top-performing systems for all three corpora. We can also note that Spotlight,  Ambiverse and Enrycher/NE, which perform poorly on AI and IITB corpora, perform much better on the MSNBC corpus.  For Spotlight and Ambiverse, this is somehow surprising: we expect SA systems to perform better on the two other corpora, which correspond to their task. 
For the F-score, Table 8 shows that TextRazor and Dandelion are the only ones found in the top-performing systems for both SA corpora (AI and IITB) despite the fact that TextRazor is not classified as a SA system.  For the NE task, results on the MSNBC corpus show that Ambiverse and Aylien/Concepts outperform the other systems.

To summarize, our results using macro-average metrics show that  despite good disambiguation capabilities, semantic annotators poorly identify entity mentions in text (spots), which leads to a weak performance for the full annotation process.  Additionally, the limited number of datasets impedes the capability to discriminate systems at a fine-grained level. In the next section, we present our results using micro-average.

\section{Evaluation using micro-average}
\label{section:evaluation:micro}

In this section,  we aggregate all spotted mentions for each system, and thus we can estimate the overall precision and recall, independently of the documents. Basically, the precision could be estimated by computing the ratio of correctly detected mentions over the total  mentions spotted by a system. Similarly, by taking the set $M$ of all mentions in the gold standard, we can estimate a recall value for each system by computing the ratio of correctly detected mentions over the number of mentions in the set $M$. The problem with this way of estimating precision and recall is that it becomes difficult to identify  statistically significant differences between systems, since we only have one ratio for each system. 

Thus we rather chose to estimate the probability that a system behaves correctly for a specific mention in a document. More precisely, to estimate the precision, we take each mention spotted by a system, and we  calculate the probability of this mention to be correct using logistic regression. Similarly, for recall, we take each mention of set $M$ and we consider the probability for the system to detect this mention. In our data, since each correctly detected mention is associated to a value 1, and 0 otherwise, we can estimate these probabilities using  logistic regression\footnote{Note that we are using logistic regression as a statistical tool, not in a machine-learning perspective. }.

\begin{table*}[htbp]
\label{precision:spotting:logit}
\centering
\caption{Estimated precision of spotting, for each system and corpus. Dash lines separate indistinguishable groups of system.}

\scriptsize
\noindent
\begin{tabular}{lllll}
\begin{tabular}{ll}
\toprule
\multicolumn{2}{c}{AI} \\
\hline
WatsonCon    & 0.92  \\
Yahoo        & 0.79  \\
\hdashline[1pt/1pt]
AylienCon    & 0.69  \\
Spotlight    & 0.63  \\
WatsonNE     & 0.57  \\
OpCalais     & 0.53  \\
Ambiverse    & 0.53  \\/
WatsonKW     & 0.51  \\
BabelNE      & 0.51  \\
MCloudEnt    & 0.48  \\
TextRazor    & 0.45  \\
AylienKW     & 0.43  \\
Dandelion    & 0.41  \\
EnrychNE     & 0.41  \\
AylienEnt    & 0.4  \\
\hdashline[1pt/1pt]
BabelCon     & 0.29  \\
EnrychKW     & 0.24  \\
Tagme        & 0.23  \\
Umbel        & 0.2  \\
MCloudKW     & 0.19  \\

\bottomrule
\end{tabular}
 & 
 \begin{tabular}{ll}
\toprule
\multicolumn{2}{c}{IITB} \\
\hline
AylienCon    & 0.75  \\
Spotlight    & 0.69  \\
\hdashline[1pt/1pt]
Dandelion    & 0.63  \\
EnrychNE     & 0.60  \\
Yahoo        & 0.52  \\
\hdashline[1pt/1pt]
Ambiverse    & 0.53  \\
TextRazor    & 0.50  \\
AylienEnt    & 0.48  \\
OpCalais     & 0.47  \\
WatsonCon    & 0.45  \\
\hdashline[1pt/1pt]
WatsonNE     & 0.45  \\
BabelNE      & 0.44  \\
MCloudEnt    & 0.44  \\
AylienKW     & 0.42  \\
MCloudKW     & 0.38  \\
BabelCon     & 0.37  \\
Umbel        & 0.36  \\
\hdashline[1pt/1pt]
Tagme        & 0.32  \\
\hdashline[1pt/1pt]
WatsonKW     & 0.21  \\
EnrychKW     & 0.15  \\
\bottomrule
\end{tabular}
 & 
\begin{tabular}{ll}
\toprule
\multicolumn{2}{c}{MSNBC} \\
\hline
Ambiverse    & 0.73  \\
EnrychNE     & 0.7  \\
AylienEnt    & 0.65  \\
AylienCon    & 0.57  \\
\hdashline[1pt/1pt]
MCloudEnt    & 0.52  \\
WatsonNE     & 0.51  \\
BabelNE      & 0.49  \\
Spotlight    & 0.46  \\
OpCalais     & 0.42  \\
Yahoo        & 0.36  \\
Dandelion    & 0.3  \\
WatsonCon    & 0.3  \\
\hdashline[1pt/1pt]
TextRazor    & 0.24  \\
WatsonKW     & 0.14  \\
AylienKW     & 0.13  \\
Tagme        & 0.085  \\
\hdashline[1pt/1pt]
BabelCon     & 0.054  \\
MCloudKW     & 0.012  \\
EnrychKW     & 0.0077  \\
Umbel        & 0.0039  \\
\bottomrule
\end{tabular}
 & 
 \begin{tabular}{ll}
\toprule
\multicolumn{2}{c}{SemEval} \\
\hline
Yahoo        & 0.26  \\
WatsonCon    & 0.14  \\
\hdashline[1pt/1pt]
WatsonKW     & 0.1  \\
AylienCon    & 0.044  \\
\hdashline[1pt/1pt]
Spotlight    & 0.03  \\
BabelCon     & 0.026  \\
OpCalais     & 0.024  \\
Dandelion    & 0.02  \\
AylienKW     & 0.019  \\
WatsonNE     & 0.017  \\
MCloudEnt    & 0.016  \\
TextRazor    & 0.014  \\
AylienEnt    & 0.013  \\
BabelNE      & 0.0099  \\
EnrychKW     & 0.0085  \\
\hdashline[1pt/1pt]
Tagme        & 0.0085  \\
Umbel        & 0.0083  \\
MCloudKW     & 0.0082  \\
Ambiverse    & 0.0079  \\
EnrychNE     & 0.0071  \\
\bottomrule
\end{tabular}
 &  
 \begin{tabular}{ll}
\toprule
\multicolumn{2}{c}{Inspec} \\
\hline
Yahoo        & 0.33  \\
AylienCon    & 0.31  \\
OpCalais     & 0.3  \\
Spotlight    & 0.24  \\
\hdashline[1pt/1pt]
WatsonKW     & 0.22  \\
MCloudEnt    & 0.15  \\
\hdashline[1pt/1pt]
Dandelion    & 0.13  \\
BabelNE      & 0.13  \\
WatsonNE     & 0.12  \\
TextRazor    & 0.12  \\
Ambiverse    & 0.12  \\
AylienEnt    & 0.11  \\
BabelCon     & 0.083  \\
\hdashline[1pt/1pt]
Tagme        & 0.076  \\
EnrychNE     & 0.071  \\
WatsonCon    & 0.067  \\
AylienKW     & 0.057  \\
\hdashline[1pt/1pt]
MCloudKW     & 0.043  \\
\hdashline[1pt/1pt]
Umbel        & 0.021  \\
\hdashline[1pt/1pt]
EnrychKW     & 0.0096  \\

\bottomrule
\end{tabular}

\\
\end{tabular}
\end{table*}%

\begin{table*}[htbp]
\label{logit:recall:spotting}
\centering
\caption{Estimated recall of spotting, for each system and corpus. Dash lines separate indistinguishable groups of system.}
\scriptsize
\noindent
\begin{tabular}{lllll}
\begin{tabular}{ll}
\toprule
\multicolumn{2}{c}{AI} \\
\hline
Dandelion    & 0.65  \\
Tagme        & 0.63  \\
TextRazor    & 0.63  \\
BabelCon     & 0.55  \\
\hdashline[1pt/1pt]
Spotlight    & 0.34  \\
AylienCon    & 0.26  \\
WatsonKW     & 0.25  \\
MCloudKW     & 0.18  \\
Umbel        & 0.12  \\
OpCalais     & 0.12  \\
\hdashline[1pt/1pt]
AylienKW     & 0.087  \\
WatsonCon    & 0.084  \\
BabelNE      & 0.083  \\
MCloudEnt    & 0.079  \\
Yahoo        & 0.079  \\
WatsonNE     & 0.07  \\
Ambiverse    & 0.056  \\
AylienEnt    & 0.039  \\
EnrychNE     & 0.032  \\
EnrychKW     & 0.017  \\
\bottomrule
\end{tabular}
 & 
 \begin{tabular}{ll}
\toprule
\multicolumn{2}{c}{IITB} \\
\hline
Tagme        & 0.72  \\
\hdashline[1pt/1pt]
BabelCon     & 0.64  \\
\hdashline[1pt/1pt]
Dandelion    & 0.49  \\
TextRazor    & 0.46  \\
\hdashline[1pt/1pt]
Spotlight    & 0.3  \\
MCloudKW     & 0.28  \\
\hdashline[1pt/1pt]
AylienCon    & 0.22  \\
\hdashline[1pt/1pt]
Umbel        & 0.17  \\
OpCalais     & 0.16  \\
BabelNE      & 0.15  \\
WatsonKW     & 0.15  \\
WatsonNE     & 0.14  \\
Ambiverse    & 0.13  \\
MCloudEnt    & 0.13  \\
AylienKW     & 0.12  \\
\hdashline[1pt/1pt]
AylienEnt    & 0.092  \\
EnrychNE     & 0.09  \\
Yahoo        & 0.071  \\
\hdashline[1pt/1pt]
WatsonCon    & 0.054  \\
\hdashline[1pt/1pt]
EnrychKW     & 0.015  \\
\bottomrule
\end{tabular}
 & 
\begin{tabular}{ll}
\toprule
\multicolumn{2}{c}{MSNBC} \\
\hline
TextRazor    & 0.83  \\
Ambiverse    & 0.76  \\
Dandelion    & 0.69  \\
BabelNE      & 0.65  \\
\hdashline[1pt/1pt]
WatsonNE     & 0.62  \\
Spotlight    & 0.61  \\
Tagme        & 0.61  \\
AylienCon    & 0.55  \\
AylienEnt    & 0.54  \\
OpCalais     & 0.54  \\
MCloudEnt    & 0.53  \\
EnrychNE     & 0.46  \\
\hdashline[1pt/1pt]
WatsonKW     & 0.33  \\
BabelCon     & 0.31  \\
Yahoo        & 0.18  \\
AylienKW     & 0.15  \\
WatsonCon    & 0.13  \\
\hdashline[1pt/1pt]
MCloudKW     & 0.031  \\
Umbel        & 0.0061  \\
EnrychKW     & 0.003  \\
\bottomrule
\end{tabular}
 & 
 \begin{tabular}{ll}
\toprule
\multicolumn{2}{c}{SemEval} \\
\hline
Dandelion    & 0.34  \\
BabelCon     & 0.33  \\
WatsonKW     & 0.31  \\
Tagme        & 0.28  \\
TextRazor    & 0.27  \\
Spotlight    & 0.19  \\
\hdashline[1pt/1pt]
Yahoo        & 0.15  \\
AylienCon    & 0.15  \\
OpCalais     & 0.11  \\
MCloudKW     & 0.087  \\
WatsonCon    & 0.073  \\
MCloudEnt    & 0.067  \\
Umbel        & 0.06  \\
\hdashline[1pt/1pt]
WatsonNE     & 0.044  \\
AylienEnt    & 0.043  \\
Ambiverse    & 0.025  \\
AylienKW     & 0.024  \\
BabelNE      & 0.013  \\
\hdashline[1pt/1pt]
EnrychNE     & 0.0077  \\
EnrychKW     & 0.0033  \\
\bottomrule
\end{tabular}
 &  
 \begin{tabular}{ll}
\toprule
\multicolumn{2}{c}{Inspec} \\
\hline
WatsonKW     & 0.46  \\
\hdashline[1pt/1pt]
Tagme        & 0.28  \\
BabelCon     & 0.25  \\
Dandelion    & 0.22  \\
Yahoo        & 0.22  \\
\hdashline[1pt/1pt]
Spotlight    & 0.12  \\
AylienKW     & 0.11  \\
AylienCon    & 0.084  \\
OpCalais     & 0.083  \\
\hdashline[1pt/1pt]
MCloudKW     & 0.057  \\
WatsonCon    & 0.044  \\
TextRazor    & 0.039  \\
\hdashline[1pt/1pt]
MCloudEnt    & 0.029  \\
BabelNE      & 0.022  \\
WatsonNE     & 0.02  \\
Umbel        & 0.019  \\
Ambiverse    & 0.017  \\
AylienEnt    & 0.016  \\
\hdashline[1pt/1pt]
EnrychNE     & 0.0082  \\
EnrychKW     & 0.007  \\
\bottomrule
\end{tabular}

\\
\end{tabular}
\end{table*}%

Tables 9 and 10 show the estimated probability values for precision and recall obtained with a logistic regression model. By comparing the results to Tables 4 and 5,  we observe that the best performing systems are globally the same ones. Note that we do not present the metrics' values computed directly on the datasets as they are similar to the logistic regression values.  

On the AI corpus, two systems  display significantly better precision than the others: Watson/Concept and Yahoo. Remember that in our macro-average evaluation, these two systems are not distinguishable from the six other systems. For recall, we get the same four best-performing systems. On IITB and MSNBC, the results are consistent with the results obtained with macro-average. 
For the KW task, we see that Yahoo is \textit{not} the only top system, in contrast to the results obtained with macro-average. In fact, Yahoo shares the top position with Watson/Concepts on SemEval  and with three other systems (Aylien/Concepts, Open Calais and Spotlight) on Inspec.

\subsection{Semantic annotators ranking}
\begin{figure*}[htbp]
\caption{Rankings based on precision and recall for spotting task, across tasks and datasets}
\begin{center}
\includegraphics[width=12cm]{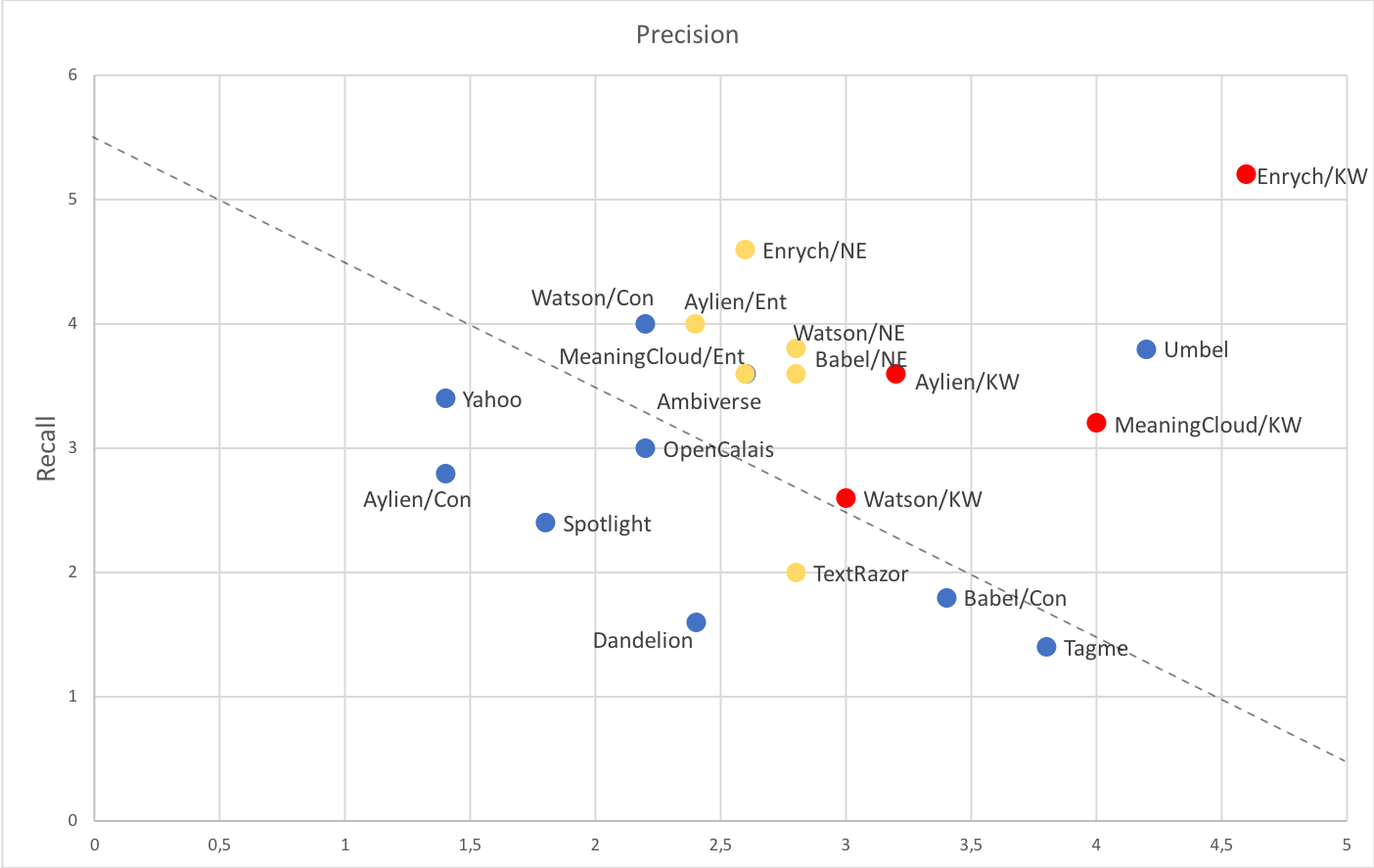}
\end{center}
\label{fig:ranking:spotting}%
\end{figure*}

Based on our previous results, we propose a global ranking score that combines the individual rankings computed separately for each dataset. These rankings are necessary to take into account groups rather than individual systems values, which are non-statistically distinguishable. Let $C$ be one of our five datasets. A system $s$ is attributed a local rank $i = Rank(s,C)$, if $s$ is the $i_{th}$  best group according to its estimated performance on dataset $C$. The global ranking score is the average of system local rank $i$ over the five datasets. Figure \ref{fig:ranking:spotting} shows  the relative position of annotators by combining their ranks for precision ($x$ axis) and recall ($y$ axis). Note that the best ranking position is 1.

If we consider precision only, Yahoo and Aylien/Concepts are the best annotators. In terms of recall, the best one is Tagme, closely followed by Dandelion. As expected, we can notice that annotators with good precision tend to have poor recall, and vice versa. 

It is interesting to note that almost all NE  and KW systems appear in the upper triangle, which means that their combined rankings in terms of precision and recall are not very good. At the opposite, most of the best systems, the ones situated in the left triangle, are SA systems.

%
%

\subsection{Logistic regression for the disambiguation task}

In this section, we repeat the same generalization process using logistic regression for the disambiguation task (Table \ref{estimated:disambiguation:precision}). Remember that what is computed here is the probability $P(\mbox{Disambiguating} |\mbox{Spotting})$, that is, we estimate the probability of correctly disambiguating the entity when a mention is correctly spotted.

We can see that on AI, the performance of the systems is not distinguishable. On IITB, Watson/Concepts significantly dominates the other annotators, and it shares this position with five other systems on MSNBC. We also note that Tagme, which does not perform very well on IITB, is among the best ones on MSNBC.

\begin{table}[htbp]
\centering
\scriptsize
\caption{Estimation of precision obtained for disambiguation task. }
\footnotesize
\begin{tabular}{p{2.5cm}p{1.2cm}p{1.2cm}p{1.2cm}}
\toprule
                   &  AI       & IITB      &  MSNBC \\
\hline
WatsonConcepts     & 1.00$^a$  & 0.95$^a$  &  0.95$^a$  \\
EnrycherEntities   & 1.00$^a$  & 0.78$^b$  &  0.69$^b$  \\
Ambiverse          & 0.91$^a$  & 0.82$^b$  &  0.80$^b$  \\
TextRazorEntities  & 0.91$^a$  & 0.76$^b$  &  0.81$^b$  \\
BabelfyNE          & 0.91$^a$  & 0.78$^b$  &  0.83$^a$  \\
Tagme              & 0.91$^a$  & 0.61$^c$  &  0.87$^a$  \\
Yahoo              & 0.89$^a$  & 0.74$^b$  &  0.76$^b$  \\
AylienConcepts     & 0.88$^a$  & 0.76$^b$  &  0.83$^a$  \\
Spotlight          & 0.86$^a$  & 0.75$^b$  &  0.82$^a$  \\
Dandelion          & 0.86$^a$  & 0.71$^c$  &  0.87$^a$  \\
BabelfyConcepts    & 0.74$^a$  & 0.55$^d$  &  0.20$^c$  \\
WatsonNE           & 0.62$^a$  & 0.59$^c$  &  0.60$^b$  \\
\bottomrule
\\
\end{tabular}%
\label{estimated:disambiguation:precision}
\end{table}%

\begin{table}[htbp]
\centering
\scriptsize
\caption{Estimation of precision and recall obtained for the full annotation process. }
\footnotesize
~\\
Precision \\
\begin{tabular}{p{2.5cm}p{1.2cm}p{1.2cm}p{1.2cm}}
\toprule
                   &  AI       & IITB      &  MSNBC \\
\hline
WatsonCon       & 0.84$^a$    & 0.23$^c$    & 0.26$^b$  \\
EnrychNE        & 0.65$^a$    & 0.46$^a$    & 0.50$^a$  \\
AylienCon       & 0.55$^a$    & 0.56$^a$    & 0.47$^a$  \\
Ambiverse       & 0.52$^b$    & 0.43$^b$    & 0.58$^a$  \\
Spotlight       & 0.50$^b$    & 0.51$^a$    & 0.37$^b$  \\
BabelNE         & 0.45$^b$    & 0.33$^c$    & 0.39$^b$  \\
Yahoo           & 0.37$^b$    & 0.39$^b$    & 0.26$^b$  \\
TextRazor       & 0.35$^b$    & 0.37$^b$    & 0.18$^c$  \\
WatsonNE        & 0.31$^b$    & 0.26$^c$    & 0.31$^b$  \\
Dandelion       & 0.30$^b$    & 0.43$^b$    & 0.24$^b$  \\
BabelCon        & 0.17$^c$    & 0.20$^d$    & 0.01$^d$  \\
Tagme           & 0.19$^c$    & 0.19$^d$    & 0.067$^d$  \\
\bottomrule
\\
\end{tabular}%

Recall \\
\begin{tabular}{p{2.5cm}p{1.2cm}p{1.2cm}p{1.2cm}}
\toprule
                   &  AI       & IITB      &  MSNBC \\
\hline
Dandelion          & 0.61$^a$  & 0.35$^a$  &  0.57$^a$  \\
Tagme              & 0.60$^a$  & 0.43$^a$  &  0.50$^a$  \\
TextRazorEntities  & 0.59$^a$  & 0.38$^a$  &  0.64$^a$  \\
BabelfyConcepts    & 0.38$^a$  & 0.35$^a$  &  0.063$^c$  \\
Spotlight          & 0.29$^b$  & 0.21$^b$  &  0.49$^a$  \\
AylienConcepts     & 0.22$^b$  & 0.16$^b$  &  0.46$^a$  \\
WatsonConcepts     & 0.079$^b$  & 0.029$^e$  &  0.13$^c$  \\
BabelfyNE          & 0.0073$^b$  & 0.11$^c$  &  0.52$^a$  \\
Ambiverse          & 0.061$^c$  & 0.10$^c$  &  0.60$^a$  \\
EnrycherEntities   & 0.055$^c$  & 0.075$^d$  &  0.33$^b$  \\
Yahoo              & 0.044$^c$  & 0.051$^d$  &  0.14$^c$  \\
WatsonNE           & 0.040$^c$  & 0.079$^c$  &  0.37$^b$  \\
\bottomrule
\\
\end{tabular}%

\label{estimated:fulltask}
\end{table}%

\subsection{Logistic regression for the full annotation process}

Table \ref{estimated:fulltask} compares the annotators on the combined task of spotting and disambiguation. Once again, we note that the obtained results are very similar to the ones with macro-average. One interesting difference is that Aylien/Concepts and Enrycher/NE are now among the top-performing systems in terms of precision for all corpora. For recall, Dandelion joined  Tagme and TextRazor in the top systems. Another clear result is that some systems display much better recall on MSNBC than on the two other corpora: Watson/Concepts, Babelfy/NE, Ambiverse, Enrycher/Entities and Watson/NE. In the case of Watson/Concepts and Ambiverse, this is somehow surprising, since they are classified as SA systems. 

\begin{figure*}[htbp]
\caption{Ranking based on precision and recall, for the full annotation process across tasks and datasets}
\begin{center}
\includegraphics[width=12cm]{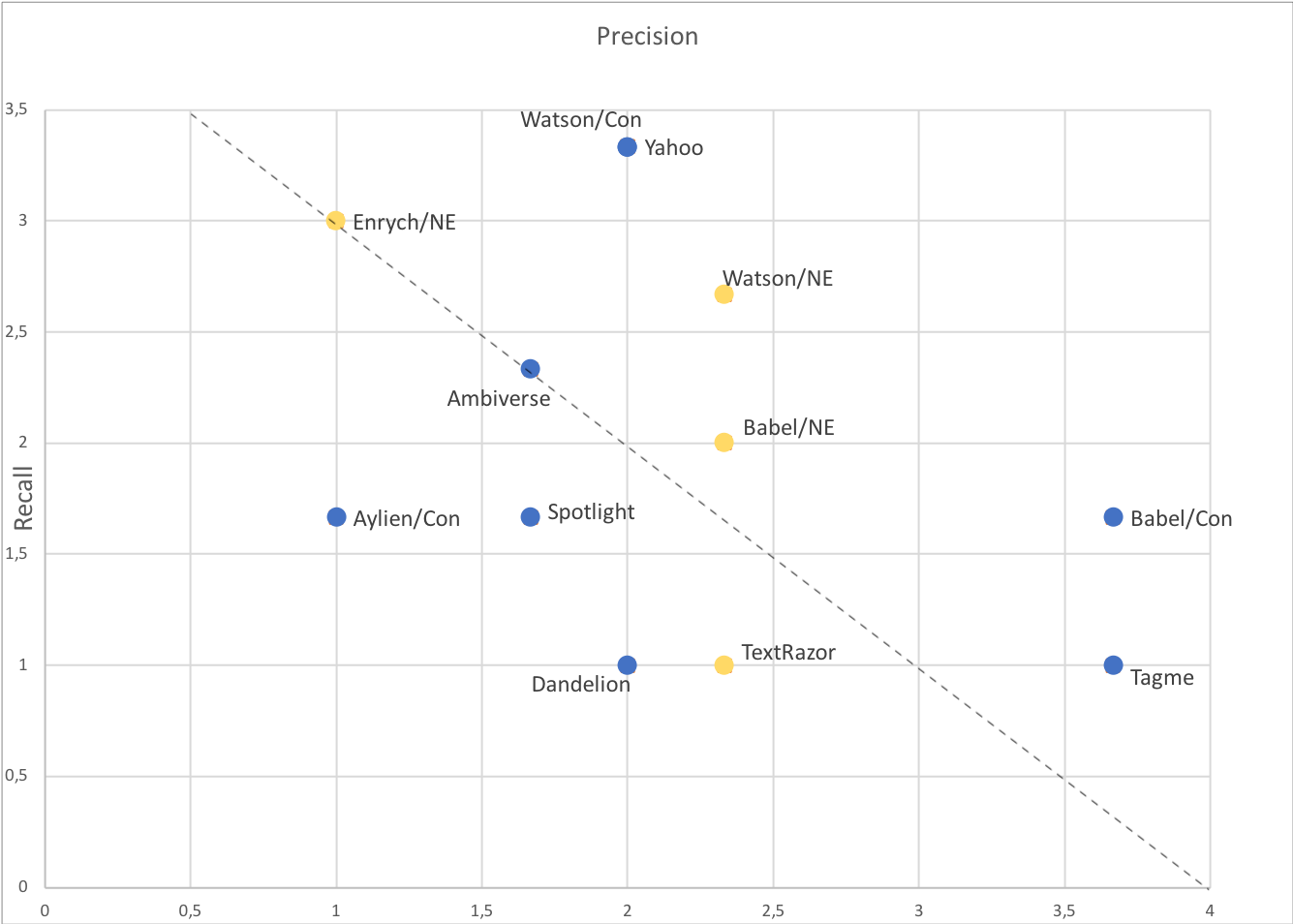}
\end{center}
\label{fig:ranking:estimated}%
\end{figure*}

Figure \ref{fig:ranking:estimated} shows the combined rankings in terms of precision and recall using the same formula as before. Overall, we can draw the same conclusion about the superiority of SA systems.

\section{Evaluation for the three tasks}
\label{section:evaluation:tasks}

%
%
%
In this section, we determine if there is some correlation between the task associated to an annotator (SA, NE or KW), and the task associated to each dataset (remember that each dataset  is also associated to one of these three tasks). We  computed a logistic regression where we consider all the systems associated to each of the three tasks. Note that we restrict here our analysis to spotting since, as we noted in our results, this is the most decisive step for the annotators' performance. Also, this enables the inclusion of KW systems in the analysis (remember that these systems do not have a disambiguation step). Results are shown in Table \ref{tab:logit:systemtype:corpus}. The first observation  is that SA annotators globally perform better than other systems on all datasets, if we consider both precision and recall. If we consider only the precision, KW systems do not perform as well as SA and NE systems. Another conclusion is that there is not any clear correspondence between the original task of a semantic annotator and the corresponding dataset. For example we cannot conclude that SA annotators perform better than others on the AI and IITB datasets, contrarily to our expectation. Similarly, NE annotators are not necessarily the best performing annotators on MSBNC and, finally, KW systems do not  dominate on the last two datasets.
\begin{table}[htbp]
\footnotesize
\centering
\caption{Estimated probabilities of correctly spotting an entity for each task and  each dataset.}

\begin{tabular}{llllll}
\\
\toprule
             &      AI      &     IITB     &    MSNBC     &   SemEval    &    Inspec     \\
\hline
\multicolumn{6}{c}{Precision} \\            
\hline
SA           &    0.3639    &    0.4267    &    0.2294    &    0.0189    &    0.1149    \\
NE           &    0.3993    &    0.4637    &    0.2570    &    0.0219    &    0.1311    \\
KW           &    0.3262    &    0.3864    &    0.2013    &    0.0161    &    0.0990    \\
SA           &    0.2795    &    0.2891    &    0.1698    &    0.2795    &    0.2795    \\
NE           &    0.3540    &    0.3648    &    0.2241    &    0.3540    &    0.3540    \\
KW           &    0.3540    &    0.3648    &    0.2241    &    0.3540    &    0.3540    \\

\bottomrule
\multicolumn{6}{c}{Recall} \\
\hline
SA           &    0.2766    &    0.2891    &    0.5202    &    0.1657    &    0.1371    \\
NE           &    0.0805    &    0.0852    &    0.1990    &    0.0435    &    0.0351    \\
KW           &    0.2287    &    0.2397    &    0.4567    &    0.1335    &    0.1097    \\

\bottomrule
\end{tabular}
\label{tab:logit:systemtype:corpus}
\end{table}

\section{Discussion and conclusion}
\label{section:discussion}

The work presented in this paper provides a comprehensive study over a wide range of semantic annotation systems for three different tasks. This kind of evaluation provides a basis for choosing a particular system depending on the task at hand, e.g. achieving a comprehensive annotation of documents, identifying named entities, or providing a small number of relevant keywords. We provide a statistical analysis of our results through ANOVA and a logistic regression model. Hereafter, we refer to our initial research question and provide some answers drawn from our experiments. \\

\noindent \textit{ How do linked data  annotators perform on the three tasks (SA, NE and KW)}?

The first major observation is the weak spotting capability of annotators across all three tasks with an F-score for the top performing systems (macro average) ranging from 0.40 to 0.53 for the SA task, from 0.17 to 0.30 for the KW task, and with only one system for the NE task reaching 0.75. 
Once the spots are identified, the majority of annotators disambiguate correctly the spots with an F-score between 0.86 and 1. However, due to the low spotting performance, the F-measure for the full annotation process only ranges from 0.31 to 0.43 for the SA task and from 0.37 to 0.47 for the NE task.

The second major observation is that semantic annotators perform better than the other two types of systems on all three tasks in terms of F-score. Even when the identified task is a keyword extraction or a named entity recognition, SA systems are still the best choice, especially AlienConcept, Yahoo, DBpedia Spotlight and Dandelion. However, the full annotation process, even for the top SA system, remains a difficult task given an F-score lower than 0.5. This is even more noticeable for the KW task, whose results do not exceed 0.30 F-score.  
When we statistically analyze tasks (SA, NE, KW) independently from the individual systems, we observe that all three system's types obtain a very low performance on the keyword extraction datasets (SemEval and Inspec). In terms of precision, SA are indistinguishable from NE (they obtain a similar results) for the SA task but they outperform NE systems in recall. Finally, SA systems outperform NE systems for the NE task.
We notice also that NE systems do not return a lot of spots and fail to identify many relevant named entities. This might be explained by the emergence of several named entity types on the linked data cloud in contrast to the more traditional named entity detection task. 
Based on our experimental results, one key insight is that semantic annotators are better at annotating all concepts in documents rather than extracting a limited set of key-phrases (keyword extraction), or relevant named entities, but are nevertheless the best performing systems for all three tasks. 

There are some limitations to our study. One improvement to our work would be to report evaluation results based on a \emph{semantic} approach. In our current work, we applied a \emph{lexical} approach for matching key-phrases and entities returned by semantic annotators with those indicated in the gold standards. As an example, based on this approach, two key-phrases 'parallel processes' and 'parallel processing' are matched because both have the same stem. This approach seems reasonable for the spotting phase. However, in the disambiguation part of the evaluation, this might cause some problems as two phrases will match only if both have been assigned to exactly the same entity in Wikipedia or any other knowledge base. This approach limits the evaluation performance by disregarding possible partial matchings that may exist between key-phrases. For example, given a gold standard phrase `parallel processing method', systems that retrieve `parallel processing', `parallel systems' or nothing will be considered as equally unsuccessful. \cite{www2013} attempt to address this issue by providing a \emph{weak annotation match} mechanism. Based on
this mechanism, two key-phrases match if they overlap and refer to the same entity in the knowledge base. However, there are situations that cannot be handled by this partial matching approach. As an example, consider one gold standard key-phrase `parallel computing', and two systems I and II that return `parallel processing' and `CPU' as key-phrases, respectively. We can assume that both systems work better than a system that returns nothing for instance, while it is obvious that System I finds a closer match to the original keyword than System II.   \cite{www2013} try to address some of these concerns by applying Milne-Witten's relatedness measure between phrases \citep{milne2013open}. Nonetheless, it does not provide a set of formalized performance measures on this basis.
To address these issues, a more comprehensive approach would have to deal with the semantics of links between entities on the linked open data cloud, such as sub-class, broader, narrower and similar links to evaluate the disambiguation performance. 
Another important limit of our evaluation is due the nature of our datasets. It is well known that the elaboration of a suitable dataset for the evaluation of semantic annotators is a very difficult task. There are usually a large number of mentions that must be annotated in a document. This can hardly be achieved automatically, and thus results in datasets that contain few documents, and makes it difficult to obtain reliable statistics. Also, by inspecting the gold standards distributed with the datasets, we note that there are usually many incorrect and missing annotations (e.g. MSNBC, IITB). Finally, in our evaluation, we used only one or two datasets for each task, which may not be sufficient to avoid biases in the analysis. Still, we do think that our results provide some interesting hints on the behaviour of semantic annotators on the three tasks.


\section{Acknowledgements}
This work was partly funded by the Royal Military College of Canada Academic Research Program and sabbatical leave fund and the NSERC discovery grant program.

\bibliographystyle{agsm}
\bibliography{biblio.bib}

\end{document}